\documentclass[runningheads]{llncs}

\usepackage{eccv}

\usepackage{eccvabbrv}

\usepackage{graphicx}
\usepackage{booktabs}

\usepackage[accsupp]{axessibility}  
\DeclareMathAlphabet{\mathbbold}{U}{bbold}{m}{n}
\usepackage[flushleft]{threeparttable}

\usepackage{amsfonts}
\usepackage{pifont}

\usepackage{hyperref}

\usepackage{orcidlink}

\begin{document}

\title{OmniColor: A Unified Framework for Multi-modal Lineart Colorization}


\titlerunning{Abbreviated paper title}

\author{Xulu Zhang\inst{1} \and
Haoqian Du\inst{1} \and
Xiaoyong Wei\inst{1,2} \and
Qing Li\inst{1}}

\authorrunning{Xulu Zhang et al.}

\institute{The Hong Kong Polytechnic University \and
Sichuan University
}

\maketitle

\begin{abstract}
  Lineart colorization is a critical stage in professional content creation, yet achieving precise and flexible results under diverse user constraints remains a significant challenge. To address this, we propose OmniColor, a unified framework for multi-modal lineart colorization that supports arbitrary combinations of control signals. Specifically, we systematically categorize guidance signals into two types: spatially-aligned conditions and semantic-reference conditions. For spatially-aligned inputs, we employ a dual-path encoding strategy paired with a Dense Feature Alignment loss to ensure rigorous boundary preservation and precise color restoration. For semantic-reference inputs, we utilize a VLM-only encoding scheme integrated with a Temporal Redundancy Elimination mechanism to filter repetitive information and enhance inference efficiency. To resolve potential input conflicts, we introduce an Adaptive Spatial-Semantic Gating module that dynamically balances multi-modal constraints. Experimental results demonstrate that OmniColor achieves superior controllability, visual quality, and temporal stability, providing a robust and practical solution for lineart colorization. The source code and dataset will be open at \hyperref[]{https://github.com/zhangxulu1996/OmniColor}.
\end{abstract}

\section{Introduction}
\label{sec:intro}





\begin{figure}[tbp]
  \centering
  \includegraphics[width=\linewidth]{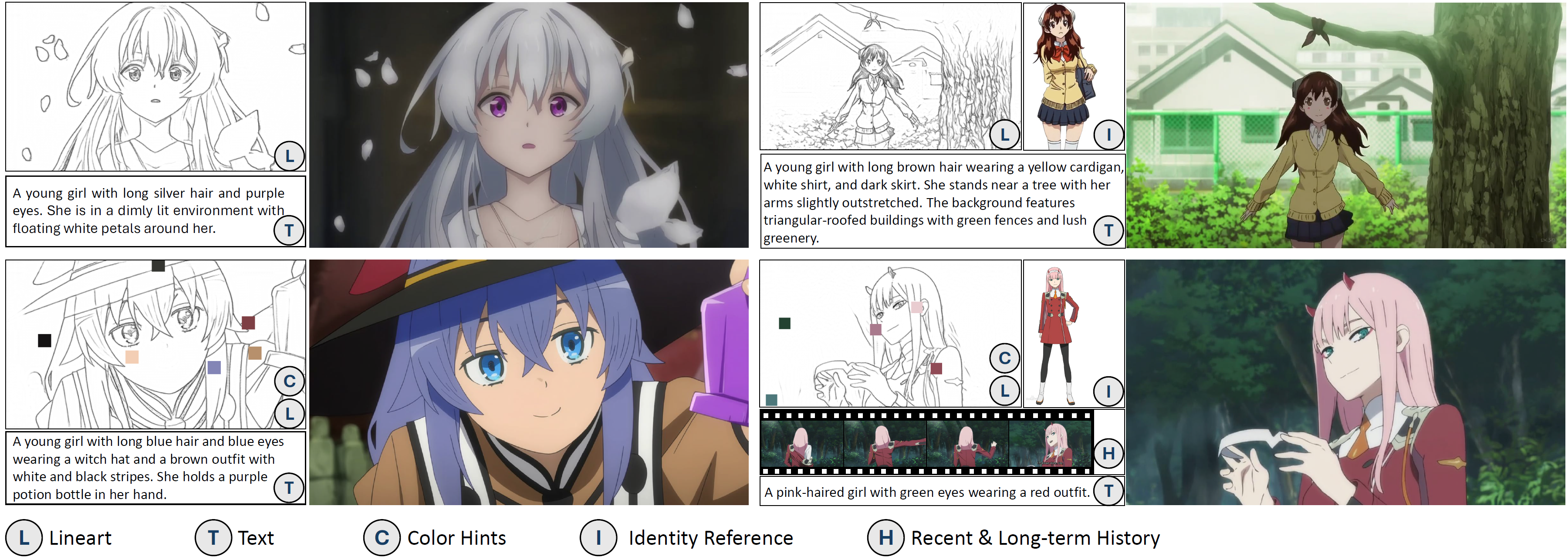}
  \caption{\textbf{Demonstration of OmniColor's unified colorization capabilities.} From a single lineart (L), our model can generate diverse, high-quality colorized results by integrating different combinations of text (T), color hints (C), identity reference (I), and temporal history (H).
  }
  \label{fig:teaser}
  \vspace{-0.3in}
\end{figure}

Lineart colorization aims to synthesize high-fidelity visual content from sparse hand-drawn structures \cite{carrillo2023diffusart,li2022eliminating,maejima2019graph}. This technology empowers a wide range of applications, including animation production, comic creation, and game design \cite{casey2021animation,huang2022unicolor,liang2025control,eva2025visual}. In this paper, we specifically focus on advancing the capabilities of controllable multi-modal lineart colorization to meet the rigorous demands of real-world creative environments.

The primary challenge in lineart colorization lies in maintaining character identity consistency and stylistic continuity across a sequence of frames \cite{liu2025manganinja,huang2024lvcd}. Relying solely on structural lineart guidance is often insufficient to capture a creator’s nuanced intent or to ensure global coherence. Consequently, integrating auxiliary control signals has become a focal point of recent research. 
Existing methods have explored various modalities such as text prompts \cite{cao2021line,kim2019tag2pix,zhang2023adding}, sparse color hints \cite{cao2021line,dou2021dual,ling2023freedrag,sangkloy2017scribbler,zhang2018two,zhang2021user}, or reference images \cite{dai2024learning,li2022style,li2022eliminating,wu2023self,wu2023flexicon,zhang2021line,liu2025manganinja,huang2024lvcd}.
However, these methods often suffer from limited flexibility. Most frameworks are designed for one additional control signal and are frequently restricted to simple scenarios, such as isolated character portraits. 
This lack of versatility hinders their application in complex, production-level scenes where multiple, potentially overlapping constraints are required. 
To bridge this gap, we present OmniColor, a unified framework that facilitates multi-modal lineart colorization through flexible combination of various control signals. This approach ensures robust colorization that meets the intricate demands of real-world creative workflows.

To effectively manage heterogeneous inputs, we categorize the guidance signals into two distinct types: spatially-aligned conditions and semantic-reference conditions. Spatially-aligned conditions, such as the input lineart, user-provided color hints, and most recent history frame provide precise pixel-level constraints on geometry and photometry. In contrast, semantic-reference conditions, including textual descriptions, long-term history frames, and identity reference maps, offer high-level guidance regarding style and character attributes without requiring exact spatial correspondence. We employ a dual-path encoding strategy to process these modalities, ensuring that the generative backbone can simultaneously respect precise structural boundaries and abstract semantic cues.
To maintain the structural integrity and the preservation of high-frequency details in spatially-aligned conditions, we employ a dual-encoder strategy that pairs a VAE \cite{kingma2013auto} for capturing precise color boundaries with a Vision-Language Model (VLM) \cite{bai2025qwen2} for localized semantic understanding. This synergy is further reinforced by a Dense Feature Alignment (DFA) loss via DINOv3 \cite{simeoni2025dinov3} supervision, which selectively refines structural correspondences during late denoising stages to ensure sharp, artifact-free results that transcend the limitations of compressed latent spaces.
For semantic-reference conditions, we utilize a VLM-only encoding scheme, leveraging compressed semantic tokens to capture essential attributes without the computational burden of dense feature maps. To further optimize efficiency in animation sequences, a Temporal Redundancy Elimination (TRE) mechanism extracts only the unique variations across frames, significantly shortening the token sequence and accelerating inference while maintaining robust long-term consistency.

Another significant hurdle in multi-condition generation is the potential for conflicting instructions. For instance, a global style reference may contradict a local color hint. To resolve such ambiguities, we introduce an Adaptive Spatial-Semantic Gating (AS-Gate) mechanism. This module dynamically modulates the influence of semantic tokens based on the local spatial context, allowing the model to prioritize explicit user guidance when necessary. This mechanism effectively prevents artifacts such as color bleeding and ensures that the model remains highly responsive to interactive editing while maintaining a coherent overall style.

Our framework demonstrates superior performance in complex colorization tasks, achieving high-fidelity results that align with both structural constraints and semantic intent, as illustrated in \cref{fig:teaser}. Through extensive experiments and user studies, we demonstrate that our method not only outperforms state-of-the-art approaches in quantitative metrics but also offers unprecedented flexibility for professional creators. By supporting diverse, mixed-modality controls, our approach provides a practical and scalable solution for the evolving needs of the digital animation and illustration industries.

\section{Related Work}

\subsection{Lineart Colorization}
Existing research on lineart colorization primarily focuses on the conditional restoration of colors and textures within the boundaries defined by hand-drawn lineart. To achieve this, modern approaches leverage diverse control signals, including text prompts \cite{cao2021line,kim2019tag2pix,zhang2023adding}, scribbles \cite{cao2021line,dou2021dual,liu2018auto,sangkloy2017scribbler,zhang2018two,zhang2021user}, and reference images \cite{meng2025anidoc,dai2024learning,li2022style,li2022eliminating,wu2023self,wu2023flexicon,zhang2021line,liu2025manganinja,yan2024colorizediffusion,zhang2025magiccolor,zhang2025animecolor}. However, text-based controls typically provide only global semantic guidance and lack the localized precision. Traditional scribble-based methods, while offering chromatic information, often fail to achieve the precise color restoration required for professional-grade results. Furthermore, reference-based approaches are limited by their dependence on dense semantic matching, which hinders the synthesis of elements absent from the reference. Moreover, existing multi-frame or video-based methods \cite{huang2024lvcd,liu2025sketchvideo} typically employ a holistic batch-processing paradigm. This rigid structure lacks scalability, as incorporating additional sketches necessitates re-executing the entire generation pipeline. In contrast, our framework supports decoupled or joint multi-modal control, enabling precise color manipulation while preserving structural integrity. By adopting a sequential generation strategy, our method ensures historical consistency while offering superior flexibility for incremental editing, backtracking, or sequence extension.

\subsection{Unified Multi-modal Generation}
Unified multi-modal generation aims to synthesize high-quality visual content by seamlessly integrating diverse input modalities within a cohesive framework. 
This requires models to possess robust multi-modal comprehension, specifically across textual and visual modalities.
To achieve this, existing research primarily follows two distinct paths. The first involves native unified models \cite{team2024chameleon,zhou2024transfusion,xie2024show,xie2025show2,wu2025janus,chen2025janus,ma2025janusflow,wang2024emu3,cui2025emu3,tong2025metamorph,deng2025emerging}, which achieve early-stage fusion of understanding and generation through large-scale end-to-end training with autoregressive or diffusion objectives. 
Alternatively, modular frameworks \cite{xia2025dreamomni2,pan2025transfer,wu2025qwen,chen2025blip3,labs2025flux,liu2025step1x,chen2025unireal,chen2025multimodal,wang2025seededit,song2025query} bridge pre-trained vision-language models \cite{bai2025qwen2,wang2024qwen2} with generative backbones \cite{esser2024scaling,peebles2023scalable,flux2024} via learnable adapters or specialized tokens.
However, all frameworks rely on laborious prompt engineering or iterative refinements to resolve ambiguities and balance semantic intent in multiple conditions. 
In contrast, we categorize the diverse conditions inherent in lineart colorization, employ customized modules for targeted processing, and design an adaptive gating mechanism, ensuring a harmonious integration of structural constraints and semantic creativity.

\begin{figure}[tb]
  \centering
  \includegraphics[width=\textwidth]{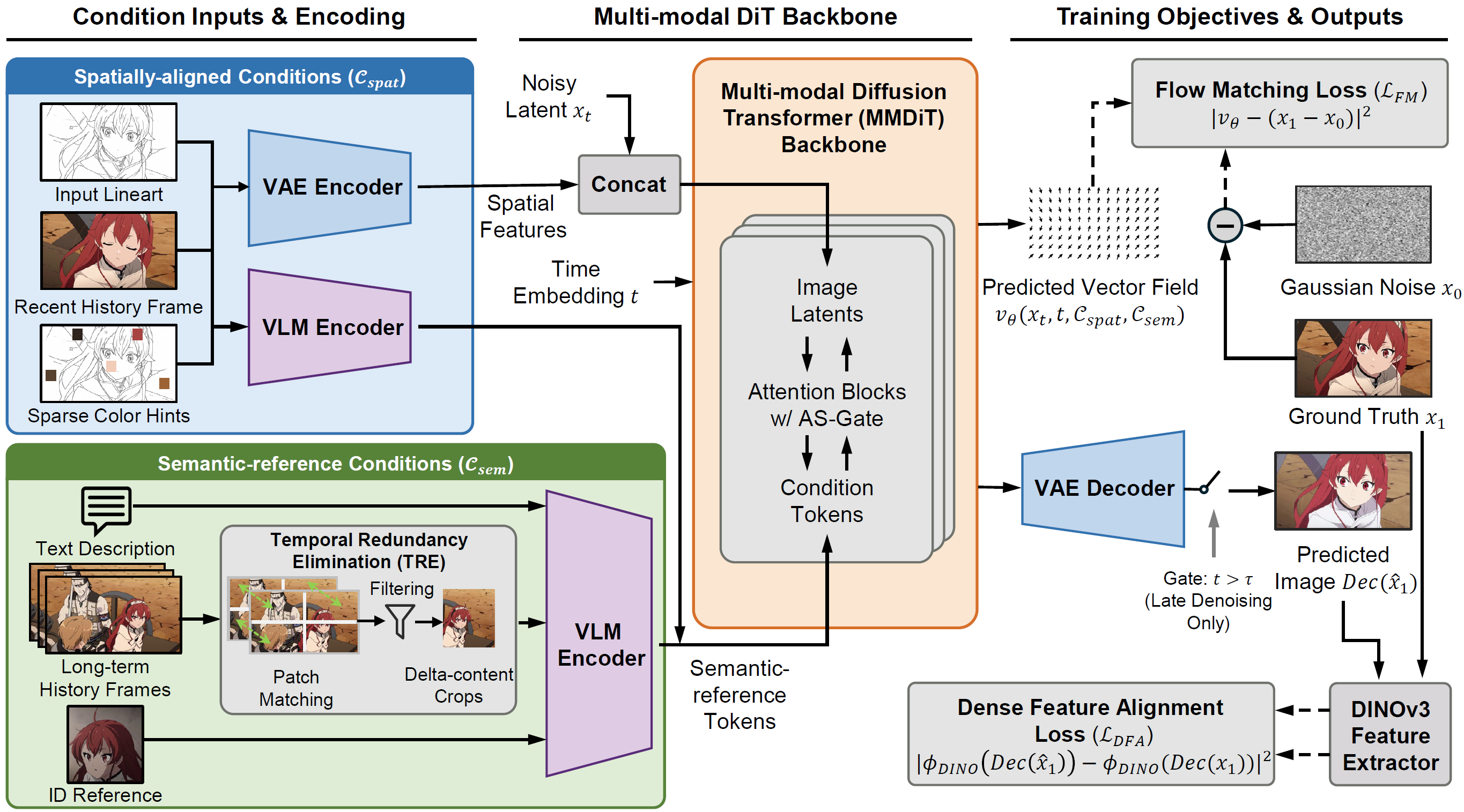}
  \caption{The architecture of OmniColor. Our framework categorizes auxiliary signals into spatially-aligned conditions ($C_{spat}$) and semantic-reference conditions ($C_{sem}$). $C_{spat}$ is processed via a dual-encoder strategy to preserve structural details. $C_{sem}$ is processed via a VLM-only encoder, where a TRE module filters redundant tokens. The MMDiT backbone integrates these features with AS-Gate module to resolve input conflicts. The model is optimized using Flow Matching Loss ($L_{FM}$) and a DFA loss.}
  \label{fig:framework}
\end{figure}


\section{Methodology}


In this section, we present OmniColor, a unified framework for controllable and consistent lineart colorization. 
As shown in \cref{fig:framework}, our architecture leverages a Multi-modal Diffusion Transformer (MMDiT) backbone \cite{peebles2023scalable} optimized via flow matching \cite{lipman2022flow} to facilitate seamless integration of diverse control signals. 
%

\subsection{Flow Matching with Multi-modal DiT}
We formulate the colorization task as a conditional flow matching problem \cite{lipman2022flow}. Let $x_1$ be the latent representation of the ground truth (GT) image and $x_0 \sim \mathcal{N}(0, \mathbf{I})$ be Gaussian noise. We define a linear probability path $x_t = (1 - t)x_0 + t x_1$, where $t \in [0, 1]$. The model $v_\theta$ is trained to predict the corresponding velocity field by minimizing the following objective:
\begin{equation}
\mathcal{L}_{FM} = \mathbb{E}_{t \sim \mathcal{U}(0,1), x_0, x_1} \left| v_\theta(x_t, t, \mathcal{C}_{spat}, \mathcal{C}_{sem}) - (x_1 - x_0) \right|^2.
\end{equation}
where $\mathcal{C}_{spat}$ and $\mathcal{C}_{sem}$ denote the spatial conditions (e.g., lineart) and semantic conditions (e.g., text), respectively. The backbone utilizes a multi-stream transformer where image latents and conditional tokens interact through joint-attention blocks, enabling bidirectional information flow between modalities.

\subsection{Spatially-aligned Condition Injection}
\textit{Definition of Spatially-aligned Conditions.}
We define $\mathcal{C}_{spat}$ as the set of conditions that require strict, pixel-level alignment with the target output. In our framework, these conditions primarily encompass the structural constraints of the input lineart, the color distribution of the most recent history frames, and user-provided sparse color hints. Unlike abstract semantic references, $\mathcal{C}{spat}$ dictates the precise geometric and photometric boundaries that the model must respect to ensure temporal stability and structural fidelity.

\textit{Dual-Encoder Feature Extraction.}
To capture the multi-scale nature of these spatially-aligned signals, we employ a dual-encoder strategy designed to extract both fine-grained textures and localized semantics. Specifically, a pre-trained VAE encoder \cite{bai2025qwen2} is utilized to preserve high-frequency details and precise color boundaries from the spatial inputs, ensuring that the model can reconstruct sharp edges. Simultaneously, a VLM encoder \cite{bai2025qwen2} extracts dense, localized semantic tokens that provide a higher-level understanding of the scene layout. By fusing these two complementary representations, the MMDiT backbone can maintain rigorous structural integrity while being aware of the underlying semantic context of the spatial constraints.

\textit{Dense Feature Alignment.}
While VAE provides an efficient latent space for generative modeling, the standard objective $\mathcal{L}_{FM}$ in this space often provides insufficient supervision to guarantee strict spatial alignment and high-frequency structural fidelity. 
To bridge this gap, we propose to perform optimization in a more dense and semantically rich feature space.
Specifically, we utilize DINOv3 \cite{simeoni2025dinov3} to extract dense features and introduce the Dense Feature Alignment (DFA) Loss to enforce this high-fidelity consistency.
Recognizing that the early stages of the flow matching process are dominated by noise, we apply this supervision selectively during the late denoising phase ($t \to 1$). 
When the decoded structures of predicted latent $\hat{x}_1$ become clear, the DFA loss minimizes the distance between the feature representations of the predicted image and the GT image:

\begin{equation}
\label{eq:dfa}
\mathcal{L}_{DFA} = \mathbbold{1}_{t > \tau} \cdot \left| \phi_{DINO}(\text{Dec}(\hat{x}_1)) - \phi_{DINO}(\text{Dec}(x_1)) \right|^2
\end{equation}

where $\text{Dec}(\cdot)$ denotes the VAE decoder, $\phi_{DINO}$ represents the DINO feature extractor, and $\tau$ is a time threshold (e.g., 0.7) that restricts the loss to the final refinement stages. This auxiliary objective encourages the model to refine fine-grained structural correspondences and accurate semantic-to-pixel mapping, leading to significantly sharper boundaries and reduced artifacts.

\subsection{Semantic-reference Condition Injection}
\label{sec:semantic}
\textit{Definition of Semantic-reference Conditions.}
We define $\mathcal{C}_{sem}$ as the set of conditions that provide high-level semantic guidance without requiring pixel-wise alignment. These conditions are inherently abstract, such as textual descriptions, character identity references, and long-term historical frames. For instance, a text prompt describing a ``moonlit night'' requires the model to generate a specific cool-toned color palette and low-key lighting, but it does not dictate the exact color values for specific coordinates.

\textit{VLM-only Semantic Encoding.} Unlike spatial conditions that necessitate a dual-encoder approach, $\mathcal{C}_{sem}$ is processed exclusively via the VLM encoder. We observe that for these abstract references, the compressed semantic tokens extracted by the VLM are sufficient to capture the necessary attributes. Besides, this VLM-only strategy offers significant computational advantages. By avoiding the generation of dense, high-resolution feature maps typically associated with VAE encoders for every reference frame, we drastically reduce the total sequence length processed by the MMDiT backbone.

\textit{Temporal Redundancy Elimination.} In animation sequences, consecutive frames often contain significant redundant information. 
To prevent the semantic branch from being overwhelmed by repetitive tokens, we propose a redundancy elimination mechanism. We decompose reference frames into patches and compute color histogram \cite{swain1992indexing} similarities between corresponding patches in adjacent frames. Patches with similarity exceeding a predefined threshold are flagged as redundant. We then perform connected component analysis on the remaining active patches. After filtering out small components, we find the minimal bounding boxes that enclose each remaining active regions. The bounding box is then used to generate a ``delta-content'' image.
This compact representation provides complementary semantic information while further shortening the token sequence, effectively accelerating inference without sacrificing consistency.

\subsection{Adaptive Spatial-Semantic Gating}
In multi-modal lineart colorization, conflicts often arise between different control signals. For instance, a dense spatial constraint (e.g., a detailed lineart) might provide sufficient structural guidance, rendering certain semantic references (e.g., text descriptions) redundant or even contradictory. Conversely, when spatial inputs are sparse, the model must rely more heavily on semantic cues. To resolve these potential conflicts and dynamically balance the contributions of different modalities, we propose the Adaptive Spatial-Semantic Gating (AS-Gate) module. Unlike static fusion methods, AS-Gate adaptively modulates the influence of the semantic branch based on the information density of the spatial branch.

The AS-Gate is integrated into the joint attention mechanism of the MMDiT blocks. Let $\mathcal{H}_{spat} \in \mathbb{R}^{L_i \times D}$ and $\mathcal{H}_{sem} \in \mathbb{R}^{L_t \times D}$ denote the hidden states of the spatially-aligned branch and the semantic-reference branch, respectively. Following the feature recalibration paradigm \cite{hu2018squeeze,peebles2023scalable}, we first derive a global spatial descriptor $\mathbf{s}_{spat}$ by pooling the image stream to capture the contextual intensity of the spatial constraints:

\begin{equation}
\mathbf{s}_{spat} = \text{MeanPool} \left( W_{up} \cdot \psi(W_{down} \cdot \mathcal{H}_{spat}) \right)
\end{equation}

where $W_{down} \in \mathbb{R}^{\frac{D}{r} \times D}$ and $W_{up} \in \mathbb{R}^{D \times \frac{D}{r}}$ form a low-rank bottleneck with reduction ratio $r$, and $\psi(\cdot)$ denotes a non-linear activation function.. This descriptor is then concatenated with the semantic features to compute the gating score $\mathbf{G}_{sem}$:

\begin{equation}
\mathcal{G}_{sem} = W'_{up} \cdot \psi(W'_{down} \cdot [\mathcal{H}_{sem}; \mathbf{s}_{spat}])
\end{equation}

The gating is restricted solely to the semantic branch, recognizing that modulating the scale of a single component is equivalent to recalibrating the relative importance between spatial and semantic constraints. The final modulated semantic output $\hat{\mathbf{H}}_{sem}$ is formulated as:

\begin{equation}
\hat{\mathcal{H}}_{sem} = \mathcal{H}_{sem} \odot \left( \sigma(\mathcal{G}_{sem}) + 0.5 \right)
\end{equation}

where $\sigma(\cdot)$ denotes the Sigmoid function and $\odot$ is the element-wise product.

To maintain training stability and preserve the pre-trained capabilities of the foundation model, we implement these projection layers using a LoRA-based structure with zero-initialization \cite{hu2022lora}. 
By initializing the up-projection weights ($W_{up}, W'_{up}$) to zero, the term $\sigma(\mathcal{G}_{sem}) + 0.5$ evaluates to $1.0$ at the start of training. 
This ensures that the AS-Gate acts as an identity mapping initially, allowing the model to gradually learn to scale the semantic contribution within a flexible range of $[0.5, 1.5]$ as training progresses.



\section{Experiments}

\subsection{Dataset}
\textbf{Training Set.} To train our unified framework, we curate a large-scale dataset featuring six types of auxiliary signals.
The raw data is sourced from 25 high-quality animation series. We extract frames with resolutions exceeding $1280 \times 720$ pixels at a frequency of 1 frame per second. We employ \textit{pySceneDetect} \cite{Castellano_PySceneDetect} for shot boundary detection to define temporal relationships.
(1) \textbf{Spatially-aligned conditions} include linearts, recent history frames, and sparse color hints.
\textbf{Lineart} is extracted using a GAN-based framework \cite{chan2022learning} specifically optimized for anime style.
\textbf{Recent history frame} is defined as the immediate preceding frame within the same shot to provide pixel-level temporal guidance. 
\textbf{Color hints} are simulated by randomly sampling uniform pixel blocks (minimum $10 \times 10$ pixels) from the ground truth.
(2) \textbf{Semantic-reference conditions} encompass text descriptions, ID reference maps, and long-term history frames.
For \textbf{textual descriptions}, we use Qwen3-VL-32B \cite{yang2025qwen3} to generate hierarchical captions (shot-level followed by frame-level) to ensure global-local consistency. 
For \textbf{ID maps}, we utilize Qwen3-VL-32B for quality filtering, SAM3 \cite{carion2025sam} for character segmentation, and CLIP features for cross-frame identity matching. 
\textbf{Long-term history frames} are sampled from adjacent but distinct shots. These frames are processed by the TRE module to remove redundant information.
Further details on data collection, annotation, and filtering are provided in \textbf{Appendix}.

The final dataset contains 120,000 high-quality pairs. During training, we adopt a stochastic sampling strategy with the following activation probabilities: Lineart (100\%), Text (95\%), Recent History (60\%), Color Hints (50\%), Long-term History (30\%), and ID Maps (15\%).

\textbf{Test Set.} The evaluation set is constructed from 6 unseen animation series. Following the same condition extraction pipeline as the training set, we obtained 305 shots and 900 test samples. Each sample is equipped with a lineart, a text description, and several color hints. Additionally, 136 samples include an ID reference image to evaluate identity-consistent generation. Furthermore, to facilitate the sequential generation workflow, we incorporate both recent and long-term history by utilizing either the ground truth or previously generated frames as inputs, where the immediate preceding frame and a distant earlier frame serve as the corresponding historical conditions.

\subsection{Experimental Setup}
\textbf{Training.} The parameters of the VAE encoder and the MMDiT backbone in our framework are initialized from Qwen-Image-Edit-2509 \cite{wu2025qwen}, and the VLM encoder adopts the pre-trained weights of Qwen2.5-VL-7B \cite{bai2025qwen2}. 
In the TRE module, we adopt a patch size of 28, set the similarity threshold to 0.85 to discard redundant patches with similarity exceeding this value, and employ a minimum connected component size of 10 to filter out small regions.
%
%
The AS-Gate employs a low-rank projection structure with a LoRA rank of $r=16$. The DFA Loss is incorporated with a weight of $\lambda_{DFA} = 0.1$ and only activated during the late denoising phase ($\tau = 0.7$ in \cref{eq:dfa}. We use the AdamW \cite{adamw} optimizer with a constant learning rate of $1 \times 10^{-5}$ and an effective batch size of 8. 
We first pre-train the MMDiT backbone for 12,000 iterations, followed by 3,000 iterations to optimize the AS-Gate while freezing the backbone on 8 NVIDIA H20 GPUs.

\textbf{Inference.} During the inference stage, the number of inference steps is set to 50 to ensure a high-quality balance between generation fidelity and computational efficiency. We adopt Classifier-Free Guidance \cite{cfg} with a scale of 4.0. All test images are generated with a resolution of $1376 \times 768$ for evaluation.

\begin{table}[ht]
    \centering
    \caption{Quantitative comparison with state-of-the-art methods.}
    \label{tab:model_comparison}
    \resizebox{\textwidth}{!}{
    \begin{tabular}{lcccccccccccc}
        \toprule
        & \multicolumn{6}{c}{\textbf{Conditions}} & \multicolumn{6}{c}{\textbf{Metrics}} \\
        \cmidrule(lr){2-7} \cmidrule(lr){8-13}
        \textbf{Method} & \textbf{L} & \textbf{T} & \textbf{H} & \textbf{C} & \textbf{G} & \textbf{I} & \textbf{Image-Align.$\uparrow$} & \textbf{SSIM $\uparrow$} & \textbf{PSNR $\uparrow$} & \textbf{FID $\downarrow$} & \textbf{LPIPS $\downarrow$} & \textbf{\boldmath $\Delta FC$ $\downarrow$} \\
        \midrule
        \multicolumn{13}{l}{\textbf{Text-based Lineart Colorization}} \\
        \midrule
        ControlNet \cite{zhang2023adding} &$\checkmark$ & $\checkmark$  & $\times$ &$\times$ & $\times$  & $\times$  &0.6403 &0.3079 &7.7390 &134.71 &0.7920 &10.13 \\
        Tag2pix \cite{kim2019tag2pix}  &$\checkmark$ & $\checkmark$  & $\times$ &$\times$ & $\times$  & $\times$ & 0.6526 &0.4525 &7.4692 &179.84 &0.5938 &3.02 \\
        \midrule
        \multicolumn{13}{l}{\textbf{Reference-based Lineart Colorization}} \\
        \midrule
        MagicColor \cite{zhang2025magiccolor} &$\checkmark$ & $\times$  & $\times$ &$\times$ & $\checkmark$  & $\times$  &0.7728 &0.5631 &10.69 &118.33 &0.5193 &8.33 \\
        AniDoc \cite{meng2025anidoc} &$\checkmark$ & $\times$  & $\times$ &$\times$ & $\checkmark$  & $\times$  & 0.8529 & 0.6638 & 17.08 & 60.47 & 0.3348 & 0.44 \\
        Manganinja \cite{liu2025manganinja}  &$\checkmark$ & $\times$  & $\times$ &$\times$ & $\checkmark$  & $\times$ & 0.9025 & 0.6914 & 16.63 & 46.86 & 0.3270 & 5.33 \\
        LVCD \cite{huang2024lvcd} &$\checkmark$ & $\times$  & $\times$ &$\times$ & $\checkmark$  & $\times$ &0.8967 &0.8220 &22.30 &36.08 &0.2699 &4.65 \\
        AnimeColor \cite{zhang2025animecolor} &$\checkmark$ & $\checkmark$  & $\times$ &$\times$ & $\checkmark$  & $\times$ &0.8376 &0.6326 &13.61 &219.47 &0.4286 &8.07 \\
        \midrule
        \multicolumn{13}{l}{\textbf{Unified Multi-modal Generation Models}} \\
        \midrule
        DreamOmni2 \cite{xia2025dreamomni2} &$\checkmark$ & $\checkmark$  & $\times$ &$\times$ & $\times$  & $\times$ &0.7273 &0.4046 &9.01 &218.95 &0.7357 &8.37 \\
        Flux-Kontext \cite{labs2025flux} &$\checkmark$ & $\checkmark$  & $\times$ &$\times$ & $\times$  & $\times$&0.7448 &0.4321 &9.20 &259.04 &0.7112 &2.92 \\
        Flux2-dev \cite{flux2} &$\checkmark$ & $\checkmark$  & $\times$ &$\times$ & $\times$  & $\times$ & 0.8061 & 0.4579 & 10.45 & 184.04 & 0.5634 & 3.61 \\
        Emu 3.5 \cite{cui2025emu3} &$\checkmark$ & $\checkmark$  & $\times$ &$\times$ & $\times$  & $\times$ &0.8144 &0.5055 &11.71 &187.85 &0.5610 &6.54 \\
        Seedream 4.5 \cite{seedream2025seedream} &$\checkmark$ & $\checkmark$  & $\times$ &$\times$ & $\times$  & $\times$ & 0.8419 & 0.6073 & 11.13 & 179.87 & 0.5996 & 7.13 \\
        Nano Banana \cite{comanici2025gemini} &$\checkmark$ & $\checkmark$  & $\times$ &$\times$ & $\times$  & $\times$ & 0.8415 & 0.5281 & 11.63 & 153.08 & 0.5466 & 8.28 \\
        Qwen-Image-Edit \cite{wu2025qwen} &$\checkmark$ & $\checkmark$  & $\times$ &$\times$ & $\times$  & $\times$ & 0.8473 & 0.5253 & 11.89 & 148.49 & 0.4682 & 8.34 \\
        \midrule
        \multicolumn{13}{l}{\textbf{Ours}} \\
        \midrule
        OmniColor  &$\checkmark$ & $\checkmark$  & $\times$ &$\times$ & $\times$  & $\times$ & 0.9242 & 0.7572 & 16.19 & 84.86 & 0.2817 & 5.44 \\
        OmniColor &$\checkmark$ & $\times$  & $\times$ &$\checkmark$ & $\times$  & $\times$ &0.9174 &0.8189 & 19.85 &86.69 & 0.2230 &6.00 \\
        OmniColor &$\checkmark$ & $\times$  & $\times$ &$\times$ & $\checkmark$  & $\times$ & 0.9749 & 0.9149 & 29.72 & 29.37 & 0.0793 & \textbf{0.09} \\
        OmniColor &$\checkmark$ & $\checkmark$  & $\times$ &$\checkmark$ & $\times$  & $\times$ & 0.9304 & 0.8121 & 19.47 & 79.64 & 0.2191 & 4.76 \\
        OmniColor &$\checkmark$ & $\checkmark$  & $\times$ &$\times$ & $\times$  & $\checkmark$ & 0.9247 & 0.7591 & 16.31 & 81.78 & 0.2541 & 5.32 \\
        OmniColor &$\checkmark$ & $\checkmark$  & $\checkmark$ &$\times$ & $\times$  & $\times$ & 0.9246 & 0.7611 &16.38 & 91.49 & 0.2785 &0.37 \\
        OmniColor &$\checkmark$ & $\checkmark$  & $\checkmark$ &$\checkmark$ & $\times$  & $\times$ &0.9310 &0.8195 &20.01 &84.97 &0.2112 &0.11 \\
        OmniColor &$\checkmark$ & $\checkmark$  & $\checkmark$ &$\checkmark$ & $\times$  & $\checkmark$ &0.9312& 0.8193 &20.03&83.92&0.2115&0.64\\
        OmniColor &$\checkmark$ & $\checkmark$  & $\checkmark$ &$\times$ & $\checkmark$  & $\times$ & \textbf{0.9769} & \textbf{0.9234} & \textbf{34.65} & \textbf{27.76} & \textbf{0.0770} & 0.14 \\
        \bottomrule
    \end{tabular}
    }
    \begin{tablenotes}[flushleft]
            \footnotesize
            \item \textbf{Note:} L=lineart; T=text description; C=color hints; I=ID reference; H=sequential generation; G=using the first GT image of each shot as reference.
        \end{tablenotes}
    \vskip -0.2in
\end{table}

\textbf{Evaluation Metrics.}
We evaluate the quality across two aspects:
\textit{1) Frame Similarity:} To assess the quality and fidelity of individual generated frames relative to the ground truth, we employ several widely-used metrics. We use \textbf{FID} \cite{heusel2017gans} to measure visual realism and \textbf{PSNR}, \textbf{LPIPS}, and \textbf{SSIM} \cite{wang2004image} to quantify pixel-level and perceptual reconstruction accuracy. Additionally, we calculate \textbf{Image-Align.} as the cosine similarity between the CLIP \cite{radford2021learning} embeddings of the generated images and their corresponding GT images to evaluate semantic alignment.
\textit{2) Frame Consistency:} To evaluate the temporal coherence across a sequence, we measure consistency following the methodology established in VBench \cite{huang2024vbench}. Specifically, for a sequence of $N$ frames, we extract features $f_i$ using a pre-trained vision backbone and calculate the average cosine similarity between consecutive frames. The frame consistency $FC$ is defined as:

\begin{equation} 
FC = \frac{1}{N-1} \sum_{i=1}^{N-1} \frac{f_i \cdot f_{i+1}}{|f_i| |f_{i+1}|} 
\end{equation} 

To quantify how accurately the generated sequence mimics the temporal dynamics of the ground truth, we define the final \textbf{Consistency Error} $\Delta FC$ as the absolute difference between the consistency of GT images $FC_{GT}$ and generated images $FC_{Gen}$: 

\begin{equation} 
\Delta FC = |FC_{GT} - FC_{Gen}| 
\end{equation} 

A smaller $\Delta FC$ indicates that the generated sequence better preserves the semantic and temporal flow inherent in the ground truth.

\subsection{Quantitative Evaluation}
\textbf{Comparison with the state-of-the-art.} We evaluate our framework against open-source lineart colorization methods published in the recent two years, covering reference-guided, text-driven, and multi-modal approaches. As reported in \cref{tab:model_comparison}, our method consistently outperforms these methods under identical input configurations. Specifically, in the L+T setting, our model surpasses all unified generation competitors by a significant margin in all metrics. This performance gain demonstrates the effectiveness of our proposed framework in bridging the gap between sparse structural inputs and high-fidelity synthesis, ensuring superior structural integrity and semantic alignment.
We compare our method with multi-modal approaches under more diverse conditions in the \textbf{Appendix}.

\textbf{Multi-condition synergy.} Our framework supports arbitrary combinations of heterogeneous control signals, and we show several representative configurations in \cref{tab:model_comparison}. A prominent observation is that combining multiple modalities consistently elevates the output quality across all metrics. For instance, the introduction of sequential generation (Condition H) substantially enhances temporal stability over basic L+T inputs, reducing the $\Delta FC$ from 5.44 to 0.37. And the further integration of sparse color hints on L+T+H conditions provides additional localized refinement, leading to enhanced rendering precision. This demonstrates the scalability of our approach for complex, production-level workflows.
Furthermore, utilizing the first frame of a shot as a spatial reference (Condition G) yields the highest fidelity, as this adjacent frame provides dense color information that effectively guides the diffusion process.

\subsection{User Study}
We conducted a user study with 10 participants (all holding a bachelor’s degree or higher) to evaluate the perceptual quality of OmniColor. Each participant evaluated the full test set under the L+T conditions. For each case, results from four models were presented in a randomized order. Participants selected the preferred result separately across three dimensions: Structural Fidelity, Instruction Following, and Visual Quality.
As shown in Table \ref{tab:user_study}, OmniColor significantly outperforms all competitors. Notably, it achieves a dominant 81.0\% preference rate in Structural Fidelity, far exceeding the second-best method (9.3\%). Our model also leads in Instruction Following (42.1\%) and Visual Quality (59.9\%), resulting in a 61.0\% overall preference. These results validate that our specialized gating and architectural design effectively bridge the gap between rigid structural constraints and high-fidelity semantic generation.


\begin{table}[htbp]
    \centering
    \caption{\textbf{Results of the User Study.} We report the average preference rates (\%) of human evaluators comparing our OmniColor against three SoTA methods across three dimensions. The best results are highlighted in \textbf{bold}.}
    \label{tab:user_study}
    \small
    \resizebox{\textwidth}{!}{
    \begin{tabular}{lcccc}
        \toprule
        \textbf{Method} & \textbf{Structural Fidelity} & \textbf{Instruction Following} & \textbf{Visual Quality} & \textbf{Overall Pref.} \\
        \midrule
        Seedream 4.5 \cite{seedream2025seedream} & 2.1 & 17.6 & 11.8 & 10.5 \\
        Nano Banana \cite{comanici2025gemini} & 7.6 & 27.8 & 15.2 & 16.9 \\
        Qwen-Image-Edit \cite{wu2025qwen} & 9.3 & 12.5 & 13.1 & 11.6 \\
        \midrule
        \textbf{OmniColor (Ours)} & \textbf{81.0} & \textbf{42.1} & \textbf{59.9} & \textbf{61.0} \\
        \bottomrule
    \end{tabular}
    }
\end{table}

\begin{figure}[htbp]
  \centering
  \includegraphics[width=\textwidth]{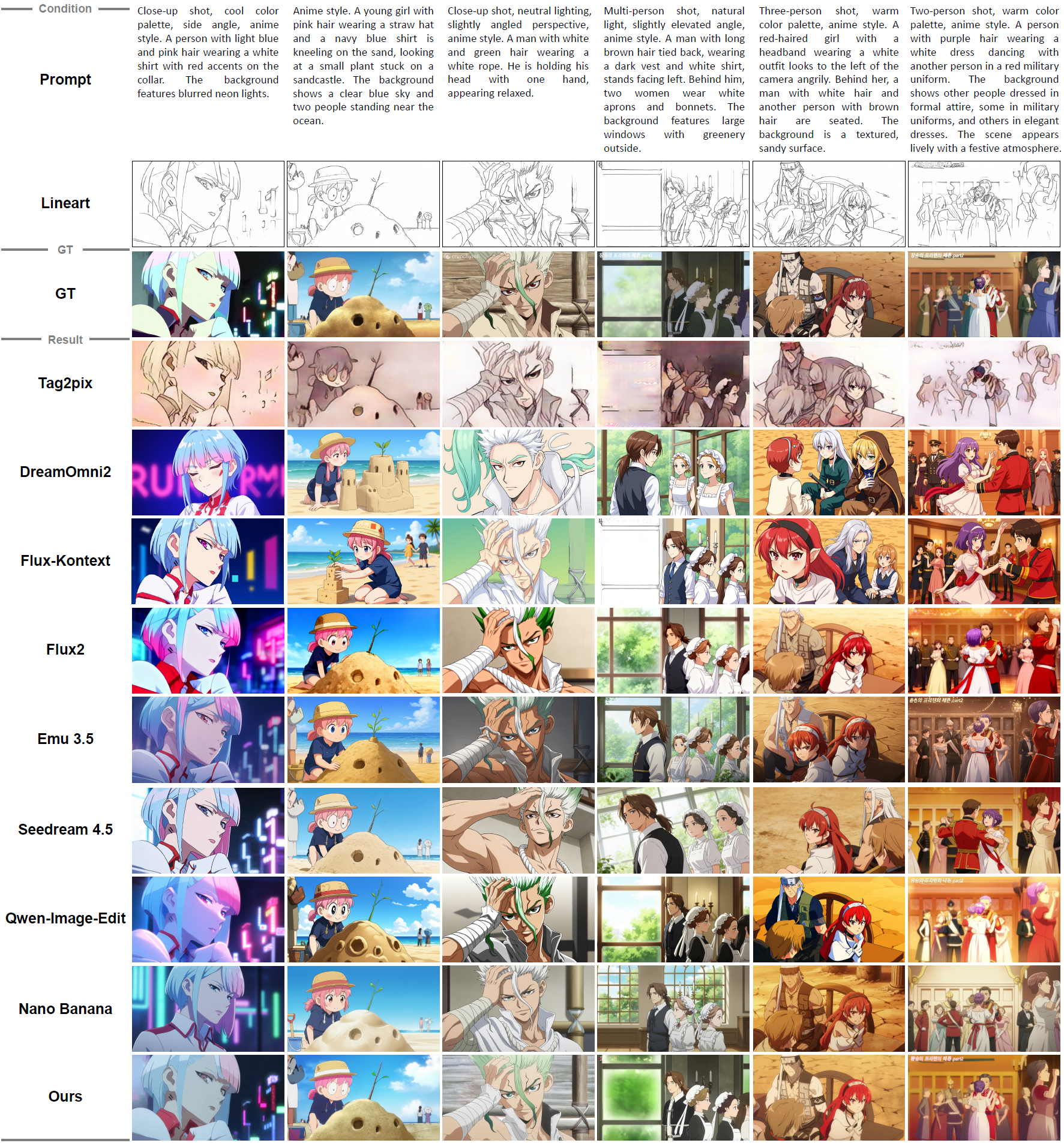}
  \caption{\textbf{Qualitative comparison of prompt-based lineart colorization.} We compare our method with Tag2pix \cite{kim2019tag2pix} and SoTA multi-modal generation models. Our framework achieves superior structural fidelity and prompt alignment.
  }
  \label{fig:sota1}
\end{figure}

\subsection{Qualitative Evaluation}

\textbf{Prompt-driven Synthesis.} We compare our method against Tag2pix \cite{kim2019tag2pix} and multi-modal generation methods in \cref{fig:sota1}. We observe that these models struggle to achieve perfect alignment with the lineart, particularly in multi-person scenes. In contrast, our approach maintains strict adherence to the input structures. Furthermore, our model demonstrates superior alignment with textual prompts, accurately rendering the colors for hair and garments and faithfully capturing the atmospheric lighting and background details described in the instructions.

\begin{figure}[htbp]
  \centering
  \includegraphics[width=\textwidth]{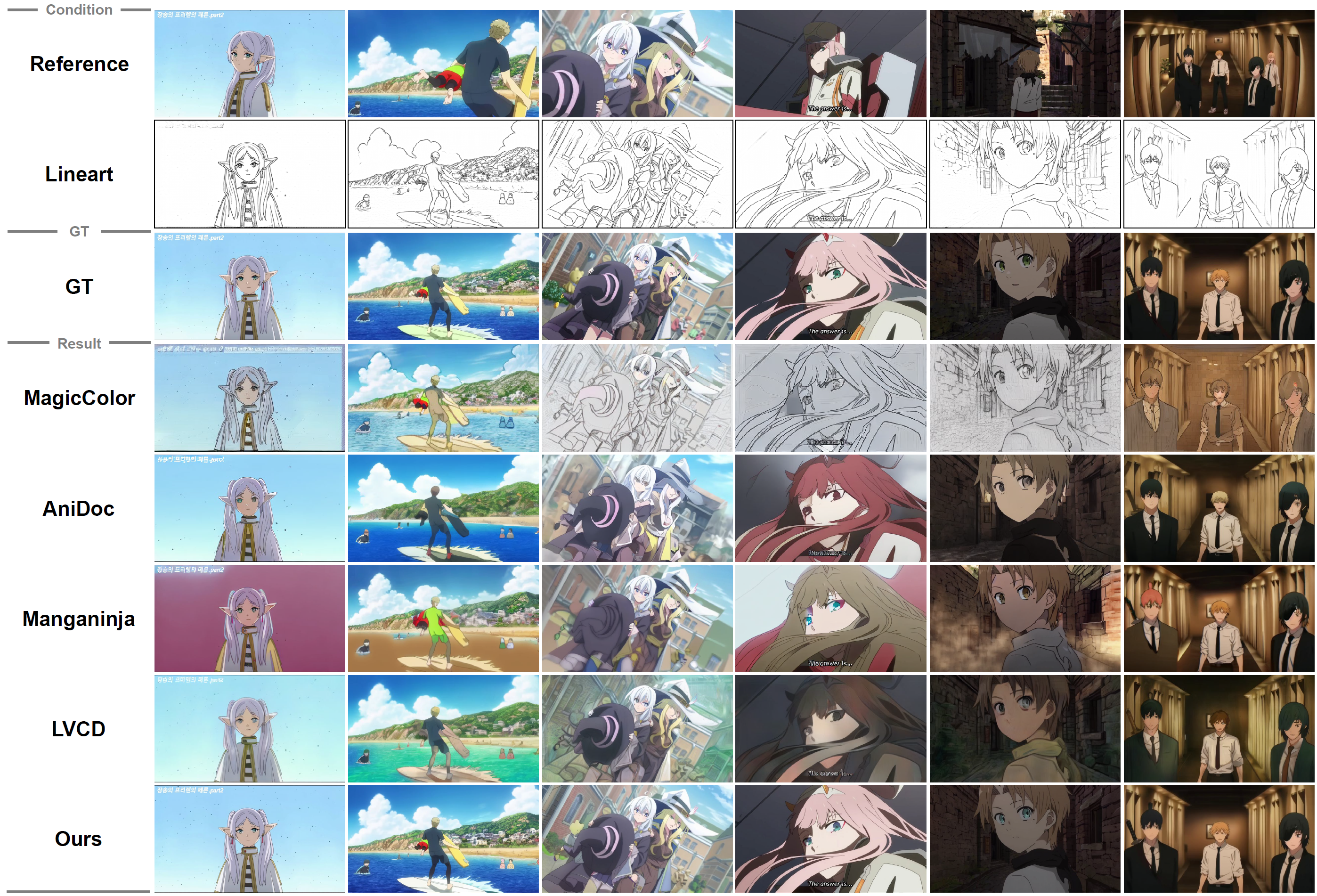}
  \caption{\textbf{Qualitative results of reference-based lineart colorization.} Our method maintains robust identity and stylistic consistency even under various motions (rotation, zoom) and significant viewpoint changes.
  }
  \label{fig:sota2}
\end{figure}

\textbf{Reference-based Colorization.} \cref{fig:sota2} showcases the performance of our framework in reference-based colorization.
Our method ensures high consistency between the reference and the target frames across diverse and challenging scenarios, including character rotations, dramatic motion, zoom-in, and zoom-out. Notably, our model exhibits remarkable stability even when the reference provides incomplete information. For instance, in cases where the reference only displays the character's back (e.g., the 4th column), our framework successfully infers and maintains the correct facial features and hair colors.

\begin{figure}[htbp]
  \centering
  \includegraphics[width=\textwidth]{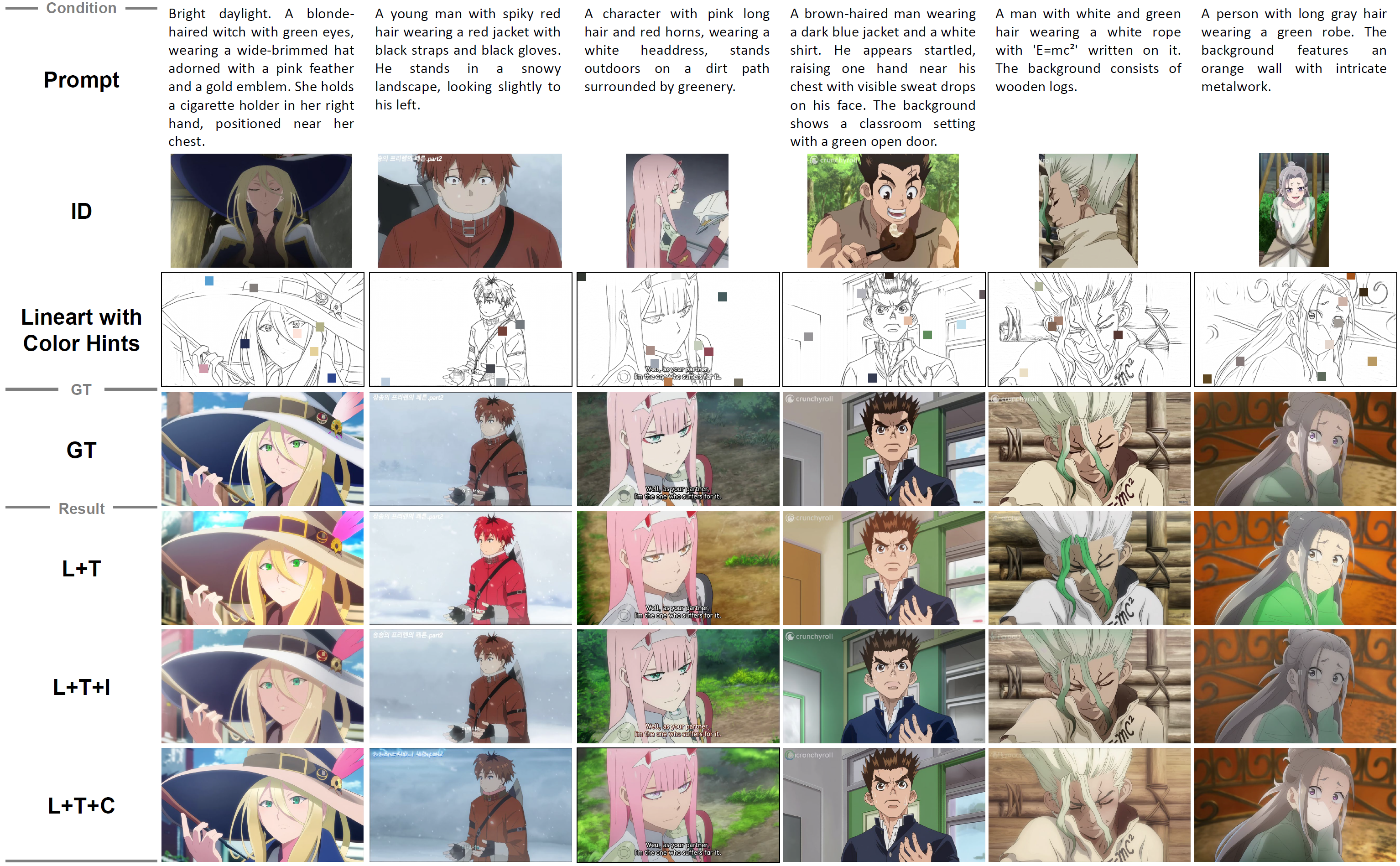}
  \caption{\textbf{Synergistic control with multiple modalities.} We show the effects of combining lineart (L) and text (T) with ID references (I) and color hints (C). The results demonstrate that our model effectively integrates multi-modal signals. 
  }
  \label{fig:color_id}
\end{figure}

\textbf{Multi-modal Control.} \cref{fig:color_id} shows the results by integrating ID maps and sparse color hints. Compared to the L+T configuration, the inclusion of an ID reference significantly enhances the precision of character-specific attributes, such as intricate makeup and hair highlights. Moreover, our framework demonstrates exceptional sensitivity to user-provided color hints. It supports a wide range of simultaneous color constraints (up to 10 distinct hints) and effectively propagates color from sparse pixel blocks to the corresponding semantic regions.

\begin{figure}[ht]
  \centering
  \includegraphics[width=0.98\textwidth]{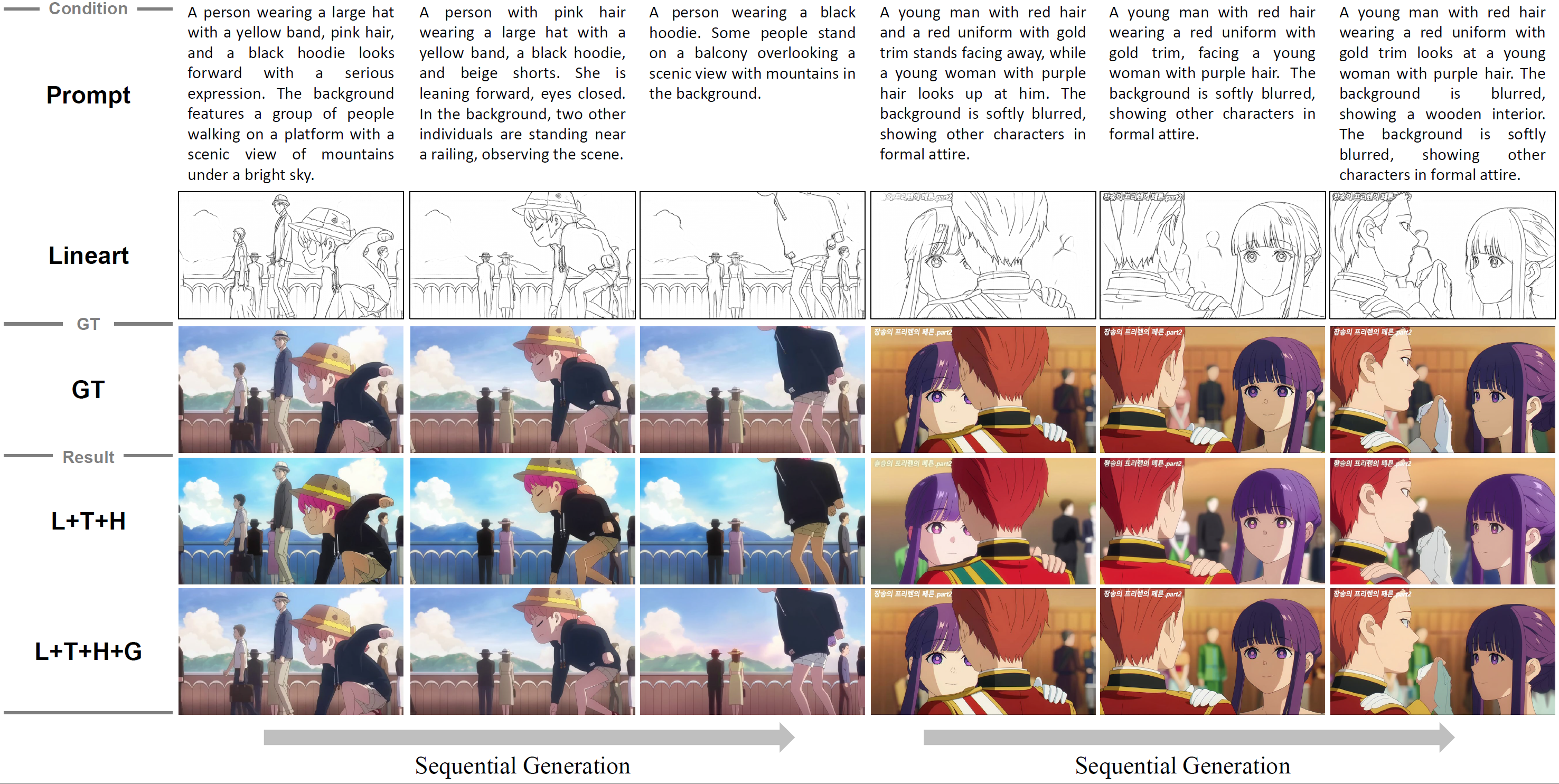}
  \caption{\textbf{Visual demonstration of sequential generation.} The sequence is generated using historical frames (H) and the 1st GT frame of the target shot (G). OmniColor ensures high temporal stability and visual coherence across the entire sequence.
  }
  \label{fig:sequential_gen}
\end{figure}

\textbf{Sequential Generation.} We demonstrate the potential of OmniColor for sequential generation. By utilizing previously generated frames as historical conditions (L+T+H), our model maintains impressive temporal continuity in character outfits, makeup, and background elements, as shown in \cref{fig:sequential_gen}. When guided by the first GT frame of a shot (L+T+H+G), the synthesized sequence achieves a visual fidelity nearly indistinguishable from the ground truth. This underscores the robustness of our iterative generation strategy in mitigating error accumulation and ensuring stable, high-quality colorization for extended sequences.

\textbf{Others.} We show the robustness across various lineart extraction techniques and evaluate the performance under conflicting conditions in \textbf{Appendix}.

\subsection{Ablation Study}
In this section, we evaluate the contribution of each key component in the L+T+H setting. 
%
%
As shown in \cref{tab:ablation_study}, adopting \textbf{VLM-only encoding} as the backbone significantly optimizes inference latency, reducing the inference time from 452s to 203s while maintaining comparable performance.
Building upon this baseline, the \textbf{DFA loss} significantly enhances structural fidelity, with SSIM increasing from 0.7287 to 0.7500 and PSNR rising by 0.79, confirming its ability to recover high-frequency details. 
%
Further incorporating the \textbf{TRE} mechanism enhances perceptual quality (FID 105.90 $\to$ 97.31) by filtering redundant history, while simultaneously boosting efficiency. It reduces history tokens from 7521 to 5203, achieving a 25\% speedup without compromising generation performance.
Finally, our full model with \textbf{AS-Gate} achieves the best performance across all quality metrics (e.g., 91.49 FID and 0.9246 Image-Align.), demonstrating that adaptive gating effectively balances spatial hints and historical semantic references. 
We show more ablation studies in \textbf{Appendix}.

\begin{table}[t]
    \centering
    \caption{\textbf{Ablation study of the proposed components.} 
    \#Tokens denotes the average token overhead of long-term history frames per sample.
    }
    \label{tab:ablation_study}
    \small 
    \resizebox{\textwidth}{!}{
    \begin{tabular}{cccccccccccc} 
        \toprule
        \multicolumn{4}{c}{\textbf{Components}} & \multicolumn{8}{c}{\textbf{Metrics}} \\
        \cmidrule(lr){1-4} \cmidrule(lr){5-12}
        \textbf{VLM-only} &\textbf{DFA} & \textbf{TRE} & \textbf{AS-Gate} & \textbf{Image-Align.$\uparrow$} & \textbf{SSIM $\uparrow$} & \textbf{PSNR $\uparrow$} & \textbf{FID $\downarrow$} & \textbf{LPIPS $\downarrow$} & \textbf{\boldmath $\Delta FC$ $\downarrow$}  &\textbf{\#Tokens $\downarrow$} &\textbf{Time (s) $\downarrow$} \\
        \midrule
        $\times$ & $\times$  & $\times$ & $\times$ & 0.9141 & 0.7371 & 15.23 & 110.32 & 0.3052 & 0.74 &  7521  & 452\\
        $\checkmark$ & $\times$  & $\times$ & $\times$ & 0.9143 & 0.7287 & 14.93 & 112.29 & 0.3143 & 0.61 &  7521  & 203\\
        $\checkmark$ & $\checkmark$ & $\times$  & $\times$ & 0.9222 & 0.7500 & 15.72 & 105.90 & 0.2943 & \textbf{0.17} &  7521 & 203\\
        $\checkmark$ & $\checkmark$ & $\checkmark$ & $\times$  & 0.9240 & 0.7588 & 16.23 & 97.31 & 0.2805 & 0.56 &  \textbf{5203} & \textbf{150}\\
        $\checkmark$ & $\checkmark$ & $\checkmark$ & $\checkmark$ & \textbf{0.9246} & \textbf{0.7611} & \textbf{16.38} & \textbf{91.49} & \textbf{0.2785} &0.37 &  \textbf{5203} & 152\\
        \bottomrule
    \end{tabular}
    }
    \vskip -0.15in
\end{table}

\section{Conclusion}
In this paper, we present OmniColor, a unified framework designed to meet the demands of multi-modal lineart colorization. By systematically categorizing control signals into spatially-aligned and semantic-reference conditions, we provide a structured approach to managing heterogeneous inputs. Our dual-path encoding strategy, reinforced by the DFA loss, ensures high-fidelity structural integrity and precise color mapping for spatial constraints. Simultaneously, the VLM-only encoding and TRE mechanism significantly optimize computational efficiency and eliminate temporal redundancies across 
We believe our framework offers a practical and scalable solution for the animation industry, promoting the way for more intuitive AI-assisted creative workflows.

\clearpage  


%
%
\bibliographystyle{splncs04}
\bibliography{main}
\end{document}